\documentclass[lettersize,journal]{IEEEtran}
\usepackage{amsmath,amsfonts}
\usepackage{algorithmic}
\usepackage{algorithm}
\usepackage{array}
\usepackage[caption=false,font=normalsize,labelfont=sf,textfont=sf]{subfig}
\usepackage{textcomp}
\usepackage{stfloats}
\usepackage{url}
\usepackage{verbatim}
\usepackage{graphicx}
\usepackage{cite}

\usepackage{hyperref}
\usepackage[dvipsnames, svgnames]{xcolor} 
\definecolor{mylightblue}{HTML}{367DBD}
\definecolor{mylightred}{HTML}{FF0000}
\hypersetup{
    colorlinks = true,   
    linkcolor = mylightred,    
    citecolor = CornflowerBlue,    
    urlcolor = blue      
}

\usepackage{amssymb}
\usepackage{booktabs} 
\usepackage{makecell}
\usepackage[normalem]{ulem} 
\usepackage{tabularx}  
\usepackage{siunitx}     
\usepackage{graphicx}    
\usepackage{multirow}
\usepackage{ragged2e} 
\AtBeginDocument{%
  }

\begin{document}

\title{HAR-DoReMi: Optimizing Data Mixture for Self-Supervised Human Activity Recognition Across Heterogeneous IMU Datasets}


\author{Lulu Ban, 
Tao Zhu,~\IEEEmembership{Senior Member,~IEEE,}
Xiangqing Lu, Qi Qiu, Wenyong Han, Shuangjian Li,
Liming Chen,~\IEEEmembership{Senior Member,~IEEE,}
Kevin I-Kai Wang,
Mingxing Nie,~\IEEEmembership{Member,~IEEE,} 
and Yaping Wan 

\thanks{This work was supported in part by the National Natural Science Foundation of China (62006110), the Natural Science Foundation of Hunan Province
(2024JJ7428, 2023JJ30518) and  the Scientific research project of Hunan Provincial Department of Education (24B0385).}

\thanks{Lulu Ban, Tao Zhu, Xiangqing Lu, Qi Qiu, Wenyong Han, Shuangjian Li Mingxing Nie and Yaping Wan are with the School of Computer Science, University of South China, 421001 China. 
(e-mail: llban@stu.usc.edu.cn, tzhu@usc.edu.cn, xqlu@stu.usc.edu.cn, qiuqi@stu.usc.edu.com, wyhan@stu.usc.edu.com, sjli@stu.usc.edu.com, niemx@usc.edu.cn, ypwan@aliyun.com) (Corresponding author: Tao Zhu)

Liming Chen is with School of Computer Science and Technology, Dalian University of Technology, 116024 China. Kevin I-Kai Wang is with Engineering and Design, Department of Electrical, Computer and Software Engineering, University of Auckland, New Zealand. (email: limingchen0922@dlut.edu.cn, email: kevin.wang@auckland.ac.nz)}}
 

\markboth{Journal of \LaTeX\ Class Files,~Vol.~14, No.~8, August~2021}%
{Shell \MakeLowercase{\textit{et al.}}: A Sample Article Using IEEEtran.cls for IEEE Journals}


\maketitle

\begin{abstract}
Cross-dataset Human Activity Recognition (HAR) suffers from limited model generalization, hindering its practical deployment. 
To address this critical challenge, inspired by the success of DoReMi in Large Language Models (LLMs), we introduce a data mixture optimization strategy for pre-training HAR models, aiming to improve the recognition performance across heterogeneous datasets.
However, directly applying DoReMi to the HAR field encounters new challenges due to the continuous, multi-channel and intrinsic heterogeneous characteristics of IMU sensor data.
To overcome these limitations, we propose a novel framework HAR-DoReMi,
which introduces a masked reconstruction task based on Mean Squared Error (MSE) loss. 
By raplacing the discrete language sequence prediction task, which relies on the Negative Log-Likelihood (NLL) loss, in the original DoReMi framework, the proposed framework is inherently more appropriate for handling the continuous and multi-channel characteristics of IMU data.
In addition, HAR-DoReMi integrates the Mahony fusion algorithm into the self-supervised HAR pre-training, aiming to mitigate the heterogeneity of varying sensor orientation. 
This is achieved by estimating the sensor orientation within each dataset and facilitating alignment with a unified coordinate system, thereby improving the cross-dataset generalization ability of the HAR model.
Experimental evaluation on multiple cross-dataset HAR transfer tasks demonstrates that HAR-DoReMi improves the accuracy by an average of 6.51\%, compared to the current state-of-the-art method with only approximately 30\% to 50\% of the data usage.
These results confirm the effectiveness of HAR-DoReMi in improving the generalization and data efficiency of pre-training HAR models, underscoring its significant potential to facilitate the practical deployment of HAR technology.

\end{abstract}


\section{Introduction}
\IEEEPARstart{T}{he} rapid growth of Internet of Things (IoT) and wearable Inertial Measurement Units (IMU) enhances the potential applications of Human Activity Recognition (HAR) in domains such as healthcare and smart homes \cite{dhekane2024transfer}. 
However, a significant challenge remains: the limited generalization ability of models across datasets, which impedes the practical deployment of HAR systems. 
Initially, most sensor-based HAR research were based on a key assumption: the training and test samples satisfy the Independent and Identically Distributed (IID) condition \cite{chen2021deep}, which is crucial to ensure good generalization performance.
However, the ubiquitous data heterogeneity in cross-dataset scenarios\cite{chang2020systematic} fundamentally undermines this assumption, resulting in a significant drop in the generalization performance of the model on unknown datasets\cite{xu2023practically}.
Moreover, publicly available HAR datasets exhibit significant non-uniformity \cite{ye2024machine} due to variations in environment, participants, and devices.
This non-uniformity consequently severely restricts the generalization of models across datasets and prevents models trained on one dataset from being effectively transferred to other datasets.

Recent cross-dataset HAR research primarily addresses these challenges through Domain Adaptation and Domain Generalization methods, often supplemented by data augmentation \cite{yan2024language} to enhance model performance and generalization ability.
Domain Adaptation \cite{chang2020systematic, qin2019cross, khan2018scaling, zhou2020xhar} seeks to improve target domain performance by mitigating source-target distribution disparity through feature alignment, fine-tuning, and adversarial training.
However, Domain Adaptation typically requires some target domain data (unlabeled or labeled), thereby reducing its practicality when target data is scarce.
Domain Generalization \cite{kim2021selfreg, saeed2019multi}, in contrast, focuses on learning domain-invariant representations to generalize to unseen target domains, obviating the requirement for the access of target domain data.
Common Domain Generalization techniques include self-supervised, contrastive, and multi-task learning. 
Data augmentation methods \cite{um2017data, xu2023practically, qian2022makes} enhance generalization by increasing training data diversity (e.g., transformations, multi-modal fusion).
However, the effectiveness of data augmentation remains limited by the inherent diversity and quality of datasets, often proving insufficient to overcome deep semantic differences between source and target domains – a primary obstacle to achieving cross-dataset generalization.

To address these limitations and combine the advantages of prior methods, self-supervised pre-training provides a promising and effective paradigm for cross-dataset generalization. 
By designing pretext tasks (e.g., masked signal reconstruction, temporal contrastive learning), these models learn transferable, general representations from large-scale unlabeled data, thereby reducing reliance on target domain labels.
These learned general features have good generalization capabilities across different tasks and domains, providing robust initialization for subsequent downstream tasks.
Notably, self-supervised pre-training has demonstrated considerable potential to enhance cross-dataset HAR performance. 
For instance, CrossHAR \cite{hong2024crosshar}, a pre-training model designed for cross-dataset HAR, leverages pre-training to learn domain-invariant features and combines data augmentation, self-supervised learning, and fine-tuning to effectively mitigate domain shift and significantly enhance performance on unseen target datasets.

Although self-supervised pre-training offers significant potential for cross-dataset HAR, existing methods often overlook the role of pre-training data composition in generalization, primarily concentrating on pretext tasks and architectures.
This leads to: \textbf{(RQ1) How can we systematically determine optimal mixture ratios during pre-training with multi-source, heterogeneous HAR datasets to maximize model generalization to unseen target datasets?} 
Inspired by the success of data mixture optimization in Large Language Models (LLMs) \cite{zhao2023survey}, particularly the effectiveness of DoReMi \cite{xie2024doremi} in enhancing LLMs generalization through optimized data mixture ratios, this paper directly addresses RQ1 by introducing data mixture optimization for pre-training HAR models. 
Specifically, we adapt DoReMi's data mixture strategy and incorporate the Group Distributionally Robust Optimization (Group DRO) \cite{sagawa2019distributionally} algorithm. 
By leveraging the Group DRO algorithm to minimize worst-domain loss, our approach optimizes the mixture ratios of heterogeneous HAR datasets, thereby enhancing the generalization of pre-training HAR models to unseen target domains.

While DoReMi has demonstrated notable success in LLMs, its direct application to HAR presents considerable adaptation challenges.
DoReMi, based on Negative Log-Likelihood (NLL) \cite{devlin2018bert} loss, is designed for discrete language sequence prediction task, which contrasts with the continuous, temporal, multi-channel nature of HAR data.
To assess the feasibility of adapting DoReMi for HAR and explore its data mixture capabilities in this domain, we ask: \textbf{(RQ2) Can the DoReMi framework be adapted for HAR while maintaining its effectiveness?} 
To address RQ2, we propose HAR-DoReMi, an innovative approach that replaces discrete language sequence prediction task with the masked reconstruction task based on Mean Squared Error (MSE) \cite{dong2015image} loss. 
This shift aims to precisely capture the dynamic patterns and numerical characteristics inherent in HAR data, thereby enhancing the effectiveness of data mixture optimization within this domain.

Beyond these challenges, IMU-based HAR data is inherently heterogeneous, further complicated by domain shifts and sensor orientation variations. 
Different sensor orientations, even at the same body placement, result in fundamentally different representations of identical movements in local coordinate systems. 
This leads to: \textbf{(RQ3) How can the inherent heterogeneity of HAR data, particularly variations in sensor orientation, be effectively addressed to train a robust HAR model across diverse datasets?} 
To address this issue, we integrate the Mahony pose fusion algorithm \cite{mahony2008nonlinear}, originally designed for pose control, into self-supervised HAR pre-training. 
By aligning multi-dataset data to a unified coordinate system, the Mahony algorithm significantly diminishes heterogeneity stemming from sensor orientation, thereby improving the cross-dataset generalization of HAR models.

To conclude, this paper introduces HAR-DoReMi, a novel pre-training framework for cross-dataset Human Activity Recognition. 
HAR-DoReMi effectively combines the Mahony pose fusion algorithm and a data mixture optimization strategy inspired by DoReMi. 
Most importantly, its HAR-domain tailored adaptation and optimization demonstrate a significant improvement in pre-training HAR model generalization on unseen target datasets compared to the original DoReMi.
Experiments on four public HAR datasets show that HAR-DoReMi achieves a significant performance boost, outperforming the state-of-the-art benchmark by approximately 6.51\% average accuracy, even with only about 30\% to 50\% of their data usage. 
The primary contributions are summarized as follows:

\begin{itemize}
\item \textbf{Data Mixture Optimization for HAR Pre-training:} 
Inspired by the success of data mixture research in LLMs, this paper pioneers the investigation of data mixture optimization for pre-training HAR models to improve the generalization on unknown datasets.

\item \textbf{DoReMi Framework Adaptation for HAR:} 
HAR-DoReMi features an innovative adaptation of the DoReMi framework for HAR, replacing its original  discrete language prediction task, with a more appropriate masked reconstruction task.

\item \textbf{Mahony Fusion Algorithm Integration for Data Heterogeneity Mitigation:} 
This paper integrates the Mahony fusion algorithm \cite{mahony2008nonlinear} into self-supervised HAR pre-training. 
The Mahony algorithm effectively mitigates the heterogeneity stemming from sensor orientation by aligning data from different datasets within a unified global coordinate system.
\end{itemize}


\section{RELATED WORK}

This section reviews related work in two key areas: cross-dataset HAR and DoReMi and the Group DRO algorithm.

\subsection{Cross-dataset Human Activity Recognition}

\begin{figure*}[ht]
\centering
\includegraphics[width=\linewidth]{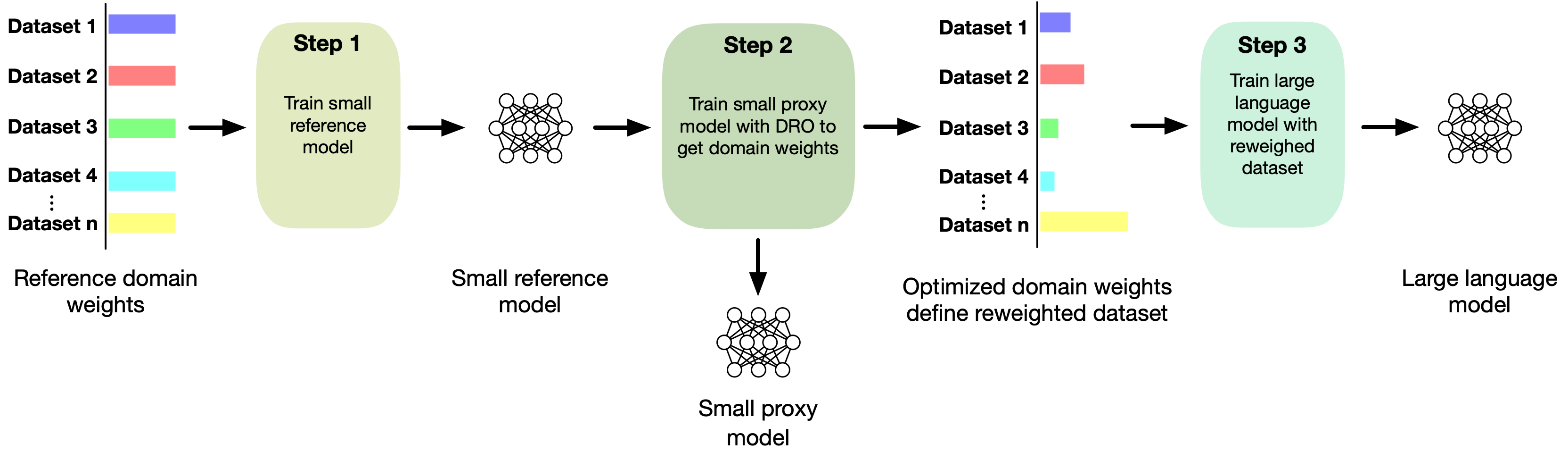}
\caption{DoReMi optimizes data mixture for large language model in three steps: (Step 1) Train a reference model with initial domain weights. (Step 2) Train a proxy model using the Group DRO algorithm to optimize domain weights. (Step 3) Train a large language model using the dataset reweighted by the optimized weights. This method effectively tunes data proportions for improving model performance.}
\label{fig_1}
\end{figure*}

IMU sensors have become the dominant method of data acquisition in HAR, owing to their portability and ease of integration.
However, the inherent heterogeneity of IMU data presents significant challenges, such as device heterogeneity \cite{stisen2015smart} (varied sensor precision), cross-modal heterogeneity \cite{li2020fusing}, dynamic data stream heterogeneity \cite{roggen2013adarc}, and deployment position variability \cite{wang2017kernel}.
Notably, the sensor settings of each HAR dataset are mostly different, which lead to large feature dimension discrepancies, significantly limiting cross-dataset generalization. 
To address these challenges, research has explored cross-dataset HAR methods, according to the degree of utilization of target domain data, broadly categorized as Domain Adaptation (DA) and Domain Generalization (DG).

A DA approach typically aims to enhance target domain performance by transferring knowledge from source domain data, leveraging both source domain data and limited amounts of target domain data(unlabeled or sparsely labeled).
In the context of HAR, DA approaches have demonstrated notable progress, primarily by focusing on reducing the divergence between source and target data distributions and learning domain-invariant feature representations.
For example, Qin et al. \cite{qin2019cross} proposed an spatial-temporal transfer learning model that can select the most appropriate source domain data and improve the model's migration performance on different datasets.
Hu et al. \cite{hu2023swl} introduced SWL-Adapt, an unsupervised DA model that bases on sample weight learning for cross-user wearable HAR. 
Mazankiewicz et al. \cite{mazankiewicz2020incremental} proposed an incremental real-time personalization approach for HAR using domain adaptive batch normalization.
Mathur et al. \cite{mathur2019unsupervised} explored unsupervised DA methods for robust sensor systems.
While these DA methods offer some domain shift mitigation, their effectiveness is strongly dependent on source and target domain similarity. 

In contrast to DA, DG methods train models solely on source data to generalize to unseen target domains without requiring the access of target domain data. 
This is often more practical, as target domain data is frequently unavailable during training. 
DG has seen increased interest in cross-dataset HAR recently. 
Presotto et al. \cite{presotto2023combining} explored combining multiple public HAR datasets to alleviate the problem of labeled data scarcity.
Inspired by the success of pseudo-labeling in semi-supervised learning, Lu et al. \cite{lu2022out} proposed pseudo-domain class labels and adversarial self-supervised pseudo-labeling methods to learn domain-invariant representations.
Qin et al. \cite{qin2022domain, qin2023generalizable} designed a series of DG methods, including a DG method based on adaptive feature fusion and a general low-resource activity recognition method based on diverse and discriminative representation learning.
Wang et al. \cite{lu2022semantic} proposed a semantic discriminative hybrid method (SDMix) to improve the generalization ability of model through data augmentation.
Kim et al. \cite{kim2021selfreg} introduced Selfreg, a self-supervised contrastive regularization approach for DG. 
Qian et al. \cite{qian2021latent} proposed a latent independent excitation method for universal cross-person activity recognition.
Xu et al. \cite{xu2023globem, xu2021limu} proposed the LIMUBERT model to learn useful IMU representations using the pre-training tasks of BERT-style.
Zhang et al. \cite{zhang2022self, zhang2023domain} proposed a self-supervised contrastive pre-training method to pre-training time series through time-frequency consistency.
Miao et al. \cite{miao2024goat} proposed the GOAT framework, a framework for universal cross-dataset activity recognition that leverages natural language supervision to improve the generalization ability of the model.

While DG methods have facilitated notable advancements in cross-dataset HAR and generalization, current approaches predominantly emphasize complex model architectures or training protocols. 
However, the composition of pre-training data is frequently overlooked, particularly in scenarios involving multiple heterogeneous datasets, where naive mixture of all datasets for pre-training can yield suboptimal results.
Addressing these limitations, this paper introduces an innovative data mixture optimization strategy specifically designed to optimize pre-training data composition, enhance the utilization of multi-source datasets, and ultimately improve model generalization to unseen target domains.

\subsection{DoReMi and the Group DRO Algorithm}



Motivated by data mixture research in LLMs, to address the challenge of multi-source dataset pre-training in heterogeneous environments, this work explores how to optimize data composition of pre-training. 
In LLMs, optimizing pre-training data composition, especially through domain re-weighting to fine-tune data source proportions, is recognized as an important way to improve training efficiency and model performance \cite{kang2024autoscale}.
DoReMi \cite{xie2024doremi} is a notable and influential work in data mixture optimization for LLMs pre-training. 
DoReMi’s core method involves training a small reference model, then using the Group DRO \cite{sagawa2019distributionally} algorithm to train a proxy model to output domain weights by minimizing excess domain loss relative to the reference model.
All models in DoReMi employ a homogeneous Decoder-only Transformer architecture, though model scales are differentiated.
Reference and proxy models are configured with 280M parameters, specifically to facilitate domain weights optimization.
Finally, a larger 8B-parameter LLM, used for downstream tasks, employs these optimized data mixture ratios to achieve high performance training.
DoReMi's architecture is illustrated in Fig. \ref{fig_1}. The underlying principles of the Group DRO algorithm are detailed below.

\begin{figure*}[ht]
\centering
\includegraphics[width=\linewidth]{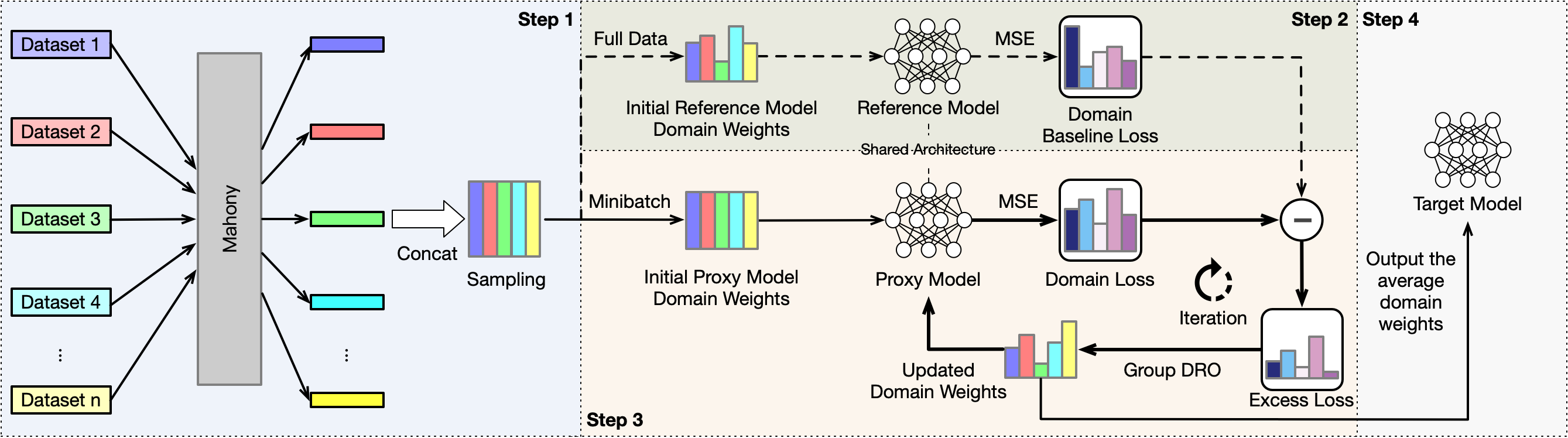}
\caption{HAR-DoReMi Framework Workflow. For scenarios without specified target datasets, the workflow comprises four steps: (Step 1) Data pre-processing via the Mahony algorithm is fed as input to model. (Step 2) A reference model is trained to establish baseline loss across each domain. (Step 3) Next, a proxy model is trained employing the Group DRO algorithm to minimize excess domain loss relative to the reference model, with outputting the average domain weights obtained by the training of proxy model. (Step 4) Lastly, training data is aggregated based on the average domain weights, and this combined data serves as input for the training of final target model.}
\label{fig_2}
\end{figure*}

The Distributionally Robust Optimization (DRO) \cite{rahimian2019distributionally} algorithm aims to enhance model generalization in the face of data distribution variations.
Distinct from the Empirical Risk Minimization (ERM) \cite{chaudhuri2011differentially}, the DRO focuses on potential data distribution uncertainty, moving beyond simply optimizing performance on the training data.
This is particularly advantageous when test data distributions diverge from training distributions \cite{sugiyama2007covariate, zhang2013domain, gama2014survey}, as it effectively mitigates distribution shift challenges.
The DRO achieves this by formulating a minimax optimization problem, aiming to find a model that minimizes the maximum potential risk across the spectrum of possible data distributions.
In essence, DRO-trained models are designed to maintain high performance even when confronted with unknown distributions from diverse datasets, thus providing a principled and robust methodology for cross-dataset generalization \cite{duchi2021learning}.
The typical formulation of DRO is the following minimax optimization problem:

\begin{equation}
\min_{\mathbf{w}\in\mathcal{W}}\sup_{\mathcal{P}\in\mathcal{S}}\{\mathbb{E}_{\mathbf{z}\sim\mathcal{P}}[\ell(\mathbf{w};\mathbf{z})]\}
\end{equation}

In this formulation, we define: $\mathcal{S}$ as the set of distributions; $\mathbf{z}\in\mathcal{Z}$ as a random sample drawn from a distribution $\mathcal{P}\in\mathcal{S}$; $\mathcal{W}$ as the hypothesis class; and $\ell(\cdot;\cdot)$ as the loss function quantifying performance \cite{zhang2023efficient}.

When $\mathcal{S}$ contains a finite number of distributions, the above problem is referred to as the Group DRO \cite{sagawa2019distributionally}. 
However, despite the DRO enhanced robustness to distribution shifts through minimax formulation, it exhibits a potential vulnerability for heterogeneous noise existing in different distributions.
This sensitivity is further amplified within the Group DRO, where disparities among candidate distributions may result in high-noise distributions unduly influencing the optimization process to obscure the contributions of other distributions.
To address this limitation, Agarwal and Zhang introduced the Minimax Regret Optimization (MRO) \cite{agarwal2022minimax} in 2022 as a refined the DRO variant. 
The MRO’s core idea is to optimize using Excess Risk, defined as the risk on each distribution minus the minimum possible risk for that distribution, instead of raw risk. 
In particular, they consider the scenario of the Group DRO, which aims to minimize the risk over multiple distributions $\mathcal{P}_1,... ,\mathcal{P}_m$ in the worst-case excess risk:



\begin{equation}
\min_{\mathbf{w}\in\mathcal{W}}\max_{i\in[m]}\big\{\underbrace{\mathbb{E}_{\mathbf{z}\sim\mathcal{P}_i}[\ell(\mathbf{w};\mathbf{z})]}_{:=R_i(\mathbf{w})}-\underbrace{\min_{\mathbf{w}\in\mathcal{W}}\mathbb{E}_{\mathbf{z}\sim\mathcal{P}_i}[\ell(\mathbf{w};\mathbf{z})]}_{:=R_i^*}\big\}
\end{equation}

This refinement can be interpreted as subtracting the intrinsic difficulty of each distribution (minimal risk $R_i^{*}$) from its raw risk, thus making the resulting excess risks more directly comparable. 
This approach not only reduces the model's sensitivity to differences in inherent noise levels in the dataset, but also makes model training more focused on the essential distribution differences between different datasets, thereby improving the model's generalization ability and robustness in data heterogeneity environments.

Despite the considerable success of data mixture optimization within LLMs, its direct application to HAR encounters significant challenges.
HAR data presents distinct the continuous and multi-channel characteristics.
To address these challenges, this paper draws on the core ideas of DoReMi and proposes an improved HAR-DoReMi method based on the characteristics of the HAR field, aiming to more effectively optimize the composition of HAR pre-training data, thereby significantly improving the generalization performance of the model in data heterogeneity environments.

\section{METHODOLOGY}


\subsection{Overall Framework for HAR-DoReMi Pre-training}

To address domain generalization challenges in cross-dataset HAR, this paper introduces HAR-DoReMi (Fig. \ref{fig_2}), an innovative pre-training framework. 
HAR-DoReMi enhances model generalization by optimizing pre-training data mixture and integrating Mahony pose fusion algorithm. 
It overcomes DoReMi's limitations in HAR and achieves superior cross-dataset generalization through four key pre-training steps: Data Pre-processing, Domain Baseline Loss Estimation, Iterative Domain Weights Updating, and Target Model Training, effectively tackling distribution shift and data heterogeneity.


In contrast to the original DoReMi framework, HAR-DoReMi introduces a key innovation: it discards the discrete language sequence prediction task based on Negative Log-Likelihood (NLL) loss and replaces it with a masked reconstruction task employing Mean Squared Error (MSE) loss, meticulously tailored to the characteristics of HAR data.
This shift to the masked reconstruction task (a self-supervised paradigm) is intended to precisely model and capture the continuous numerical features, inherent temporal dependencies, and multi-channel interrelations of HAR data, moving beyond the limitations of discrete probability distribution prediction. 
By randomly masking segments of the time series data, the masked reconstruction task effectively compels the model to learn dynamic patterns and latent representations inherent in HAR data, ensuring the extraction of more generalizable representations from relatively sparse sensor inputs. 
This shift empowers HAR-DoReMi to offer general representation learning for HAR data, even in the absence of specified downstream tasks, thereby establishing a solid foundation for domain weights optimization and cross-dataset generalization.


\subsection{Data Pre-processing Based on Mahony Algorithm (Step 1)}

A key challenge in cross-dataset HAR stems from data heterogeneity, particularly variations in sensor device orientation.
Even when sensors are deployed on identical body parts, different acquisition orientations lead to markedly distinct IMU data patterns within local coordinate systems for identical physical movements. 
This local coordinate system bias significantly restricts cross-dataset generalization, hindering effective model transfer between heterogeneous datasets. 
To address this critical issue and inspired by UniMTS \cite{zhang2024unimts}, this paper innovatively incorporates the Mahony pose fusion algorithm \cite{mahony2008nonlinear} into the data pre-processing pipeline. 
The algorithm's primary goal is to transform sensor data from varying local coordinate systems to a consistent global coordinate system, effectively mitigating data heterogeneity arising from the differences of device deployment orientation, and thus providing a more homogeneous data foundation for subsequent model training.

The Mahony algorithm is a computationally efficient complementary filter that accurately estimates sensor pose (orientation) by fusing gravity vector data from accelerometers and angular velocity data from gyroscopes. 
Distinguished by its low computational complexity, ease of implementation, and robust performance in dynamic conditions, the Mahony algorithm is ideally suited for real-time HAR.
The North-East-Down (NED) coordinate system \cite{cai2011coordinate} is employed in this work as the unified global coordinate system.
With its origin at the IMU centroid, the NED system establishes a fixed coordinate system, defining the x-axis as true north, the y-axis as true east, and the z-axis aligned with gravity. 
Consequently, this pose-independent and fixed NED framework enables the uniform mapping of IMU data from diverse datasets into a unified coordinate system, effectively eliminating signal variations arising from differences in sensor deployment orientation.

Specifically, for each timestep $t$ from $1$ to $T$ ($t\in\{1,2,........ ,T\}$) in every dataset, we initialize the orientation quaternion as $q_0=[1,0,0,0]$. 
This quaternion signifies the IMU’s initial alignment with the NED coordinate system, where the IMU’s local coordinate system overlaps the NED system. 
Subsequently, the Mahony algorithm is provided with the IMU data for the current timestep $t$. 
This input data comprises 6-axis or 9-axis measurements, including accelerometer and gyroscope data, and optionally magnetometer data.
The algorithm outputs an updated quaternion $q_t=[q_w,q_x,q_y,q_z]$, accurately representing the IMU’s orientation pose relative to the NED coordinate system at timestep $t$. 
Based on this quaternion, we can derive the rotation matrix $M_\mathrm{t}$ from the local to the global coordinate system.

\begin{equation}
\small
M_{\mathrm{t}}=
\begin{bmatrix}
1-2q_y^2-2q_z^2 & 2q_xq_y-2q_wq_z & 2q_xq_z+2q_wq_y \\
2q_xq_y+2q_wq_z & 1-2q_x^2-2q_z^2 & 2q_yq_z-2q_wq_x \\
2q_xq_z-2q_wq_y & 2q_yq_z+2q_wq_x & 1-2q_x^2-2q_y^2
\end{bmatrix}
\end{equation}

Using the rotation matrix $M_\mathrm{t}$, we can accurately transform the IMU data $X_t$ at timestep $t$ from the local to the global coordinate system, yielding $\tilde{X}_{\mathrm{t}}{:}$

\begin{equation}
\tilde{X}_\mathrm{t}=M_\mathrm{t}\cdot{X}_\mathrm{t}
\end{equation}

Following these steps, IMU data from all datasets is transformed into the unified NED coordinate system, effectively eliminating data heterogeneity arising from varied device orientations. 
Then, the data processed by Mahony algorithm is used as input for subsequent model training, establishing a basis for domain-invariant feature representation learning and ultimately improving model cross-dataset generalization.

\subsection{Domain Baseline Loss Estimation Based on Reference Model (Step 2)}

After the Data Pre-processing of Step 1, the HAR-DoReMi framework proceeds to the Domain Baseline Loss Estimation stage. 
The purpose of this stage is to train a reference model—sharing the same network architecture with the proxy model—utilizing the pre-processed HAR data from Step 1.
For reference model training, all datasets, pre-processed by the Mahony algorithm, are merged to precisely capture and quantify the Domain Baseline Loss for each data domain.

The Initial Reference Model Domain Weights represent the initial mixture ratios of datasets in the reference model training. 
To ensure the reference model effectively learns and represents the feature distribution of each data domain, we initialize the domain weights based on dataset size. 
Specifically, we determine the initial domain weights by calculating the proportion of each dataset's sample count relative to the total sample count across all datasets, directly assigning this proportion as the initial weight for each dataset in the reference model training.
This initialization, proportional to dataset size, aims to establish a stable, representative baseline for subsequent proxy model training, thereby facilitating comparable and effective excess loss calculation.
Crucially, the reference model serves solely as a baseline provider within this framework and its domain weights are fixed throughout training to ensure objective and stable baseline assessment.

\subsection{Iterative Domain Weights Updating Based on Proxy Model (Step 3)}

Following the Domain Baseline Loss Estimation stage (Step 2), the HAR-DoReMi framework transitions to the Iterative Domain Weights Updating phase. 
In this stage, the data pre-processed by the Mahony algorithm in Step 1 is used as input. 
According to the preset initial domain weights ratio of the proxy model (usually set to uniform distribution), a data subset is extracted in minibatch units for iterative training of the proxy model, which shares the same architecture as the reference model.
Unlike the reference model, however, the training data composition of proxy model is dynamically adjusted via the Group DRO algorithm to iteratively optimize domain weights.
The Initial Proxy Model Domain Weights serve to represent the initial mixture ratios of datasets within the proxy model training process.
To ensure impartial initial assessment of each dataset's contribution for generalization and mitigate potential bias from initial weight assignments on final optimization, HAR-DoReMi initializes proxy model domain weights to a uniform distribution.

After each time the proxy model is trained in minibatch units, the HAR-DoReMi framework dynamically updates the domain weights of training data with the Group DRO algorithm. 
The Group DRO algorithm aims to minimize the loss of the worst-performing domain, thereby enhancing model robustness and generalization across all domains.
The implementation procedure involves first calculating the proxy model's current Domain Loss for each domain, then computing Excess Loss as the difference between the current Domain Loss and the Domain Baseline Loss from Step 2.
Excess Loss measures the change in proxy model performance compared to the baseline for each domain, reflecting domain learning difficulty and generalization potential.
HAR-DoReMi then uses the Group DRO algorithm to iteratively update each domain's weight based on the Excess Loss.
Domains with higher Excess Loss – indicating less performance improvement or even performance decline – receive higher weights, guiding the proxy model to prioritize performance gains in these domains in subsequent training iterations. 
This iterative updating mechanism progressively optimizes the training data composition of proxy model, aiming to effectively balance learning difficulty across domains and ultimately improve model cross-dataset generalization.

Upon completion of iterative training, the HAR-DoReMi framework outputs the average domain weights from proxy model training as the final optimized domain weights, which are then used in subsequent target model training.

\subsection{Target Model Training Based on Optimized Data Recombination (Step 4)}

With the Iterative Domain Weights Updating completed in Step 3, the HAR-DoReMi framework proceeds to the final Target Model Training stage. 
The central task of this stage is to optimally mix the pre-processed training data, guided by the average domain weights derived in Step 3.
Specifically, we first divide each dataset's size by its corresponding weight from Step 3, take the minimum as the total training sample size, and then randomly sample and combine data from each dataset based on this size and their weights.
This data mixture optimization strategy enables HAR-DoReMi to effectively mitigate the negative effects of varying data distributions across datasets, ultimately yielding a more robust and generalizable cross-dataset HAR target model, and maximizing its cross-dataset generalization performance on unseen target datasets.

\subsection{The Algorithm of Step 3}

Suppose that we have k HAR domains (e.g., HHAR, UCI), with each domain $i$ containing a set of examples $D_i$.
Domain weights $\alpha\in\Delta^k$ specify a probability distribution over the k domains, and consequently a distribution over the training data: $P_\alpha=\Sigma_{i=1}^k\alpha_i\cdot\mathcal{N}$ where $\mathcal{N}$ is the total numbers of training data. 
The general process of domain weights updating is summarized in Algorithm \ref{alg1}.

\begin{algorithm}
\caption{HAR-DoReMi Domain Reweighting (Step 3)}
\label{alg1}
\begin{algorithmic}
\REQUIRE Domain data $D_1, \dots, D_k$, batch size $b$, step size $\eta$,\\
        number of training steps $T$,  smoothing parameter $c \in [0,1]$, \\ 
        distribution over the training data $P_{\alpha_{t-1}}$, 
        the indicator function $\mathbb{I}(\cdot)$.
\STATE Initialize proxy model $\theta_0$
\STATE Initialize proxy model domain weights $\alpha_0=\frac{1}{k}\mathbf{1}$
\FOR{$t = 1$ to $T$}
    \STATE Sample minibatch $B = \{x_1, \dots, x_b\}$ from $P_{\alpha_{t-1}}$, where samples in $B$ are from different domains  
    \STATE Let $|x|$ be the number of samples in $B$ from the same domain $i$, where $x\in B\cap D_{i}$
    \STATE Compute per-domain excess losses for each domain $i \in \{1,2,\dots, k\}$, where $\ell_{ref,i}(x)$ is the baseline loss of the reference model in domain $i$ ($\ell_{\theta,j}(x)$ and $\ell_{ref,i}(x)$ both are sample-level loss):
    \STATE \[\textstyle \lambda_t[i] \gets 
    \frac{\sum_{j=1}^{|x|}\max\{\ell_{\theta_{t-1},j}(x)-\ell_{ref,i}(x),0\}}{\sum_{j=1}^{|x|}\mathbb{I}(\ell_{\theta_{t-1},j}(x)-\ell_{ref,i}(x)>0)} \]
    \STATE Update domain weights (exp is entrywise):  
    \STATE \[\textstyle \alpha_{t}^{\prime}\leftarrow\alpha_{t-1}\exp{(\eta\lambda_{t})}\]
    \STATE Renormalize and smooth domain weights: 
    \STATE \[\textstyle \alpha_{t}\leftarrow(1-c)\frac{\alpha_{t}^{\prime}}{\sum_{i=1}^{k}\alpha_{t}^{\prime}[i]}+c\frac{1}{k} \]
    \STATE Update proxy model $\theta_t$ and distribution over the training data $P_{\alpha_t}$
\ENDFOR
\RETURN $\frac{1}{T} \sum_{t=1}^{T} \alpha_{t}$
\end{algorithmic}
\end{algorithm}

\section{EXPERIMENTS}

\subsection{Experiment Setup}

\subsubsection{Datasets}

Four publicly available HAR datasets were chosen for this study. 
These datasets are not only widely adopted in research \cite{xu2021limu,yang2015deep, yao2017deepsense}, but also exhibit rich diversity across device types, activities, user groups, and environments. 
This diversity makes these datasets ideal benchmarks for comprehensively evaluating model performance and comparing against current baselines.
Detailed statistics of these datasets are summarized in Table \ref{tab1}.

\begin{table*}[ht]
\centering
\small
\caption{Datasets summary (A: accelerometer, G: gyroscope, M: magnetometer and Motion actually represents MotionSense.)}
\label{tab1}
\renewcommand{\arraystretch}{1.2} 
\begin{tabular}{l >{\raggedright\arraybackslash}p{1.2cm} ccccc} 
    \toprule
    Dataset & Sensor & Activity & Users & \makecell{Placement} & Sampling Rate (Hz) & Samples \\
    \midrule
    HHAR \cite{stisen2015smart}& A, G & 6 & 9 & -- & $50\sim200$ & 7968 \\
    Motion \cite{malekzadeh2019mobile}& A, G & 6 & 24 & Front pocket & 50 & 4108 \\
    Shoaib \cite{shoaib2014fusion}& A, G, M & 7 & 10 & \makecell{Left pocket, right pocket,\\ wrist, upper arm, belt} & 50 & 7500 \\
    UCI \cite{reyes2016transition}& A, G & 6 & 30 & Waist & 50 & 1687 \\
    \bottomrule
\end{tabular}
\end{table*}

\subsubsection{Data Pre-processing}

This study employs a uniform data pre-processing pipeline to ensure consistency and comparability across datasets. 
For the four public HAR datasets, we extracted common six-channel sensor data: tri-axial accelerometer and tri-axial gyroscope.

To simulate resource-limited deployment, and consistent with prior work \cite{xu2021limu,xu2023practically}, sensor data was downsampled to 20Hz.
Data segmentation was then performed using non-overlapping sliding windows of length 120 consecutive IMU measurements.
Each window was labeled with activity class and device location (if available). 
For cross-dataset generalization validation, we evaluated performance on four common activities across all datasets: Still, Walking, Upstairs, and Downstairs. 
For Still activity, semantically similar activity classes were unified across datasets, for instance, \textquotedblleft{}Sitting\textquotedblright{} and \textquotedblleft{}Standing\textquotedblright{} were combined into the \textquotedblleft{}Still\textquotedblright{} class in HHAR.
Regardless of the specific model—be it the reference model, the proxy model, or the final CrossHAR \cite{hong2024crosshar} target model —all models uniformly utilized the data resulting from the pre-processing steps detailed above as their input.

\subsubsection{Baselines and Evaluation Metric}

For comprehensive performance evaluation and comparative analysis, we select a set of representative deep learning and state-of-the-art time series classification models as baselines, covering a wide range of model architectures.
These included: MLP \cite{ismail2019deep}, CNN \cite{gamboa2017deep}, ResNet \cite{he2016deep}, LSTM \cite{siami2019performance}, TSFCN \cite{wang2017time}, LIMUBERT \cite{xu2021limu}, ContraTSC \cite{eldele2023self}, and SDMix \cite{lu2022semantic}. 
Since our final target model architecture is CrossHAR \cite{hong2024crosshar}, we also include it in the baseline comparison to evaluate the performance improvement of our method over this state-of-the-art method.

For the evaluation metric, we use the average accuracy \cite{li2024harmamba} to evaluate the performance of the model on the multi-class human activity classification task.
The average accuracy is calculated as the average of the classification accuracy of each class, which effectively addresses the impact of class imbalance in the dataset and provides a fair and reliable performance measure for our method.
Using the average accuracy to compare with these widely used baseline models helps to fully understand the performance of our method in the human activity recognition task.

 \subsubsection{Implementation Details}

The original DoReMi framework implementation includes three model types: a reference model, a proxy model, and a final, large language model. 
The architecture of HAR-DoReMi is inspired by the core idea of DoReMi \cite{xie2024doremi}, but is specifically tuned for human activity HAR needs.
It leverages the masked reconstruction task (which is more consistent with HAR data) to significantly improve the domain generalization ability in cross-dataset HAR.

Specifically, both the reference model and the proxy model of HAR-DoReMi adopt an enhanced masked signal reconstruction model combined with the temporal channel masking \cite{wang2024improved} strategy to comprehensively capture multi-dimensional time series features. 
In order to balance training efficiency and domain weights optimization, we set the channel masking number to 3 and the masking rate of the time dimension to 70\%, reducing the amount of computation while maintaining performance.
For the stability of domain weights optimization, the training epochs of the reference model are set to 200 for sufficient baseline loss estimation, while the proxy model iteratively optimizes the domain weights in minibatch for 1000 steps.
The model architecture uses a 3-layer Transformer Encoder layer as the Encoder, and the Decoder uses a fully connected layer. 
The fully connected layer uses SwiGLU \cite{shazeer2020glu} as the activation function and RMSNorm \cite{zhang2019root} as the normalization method. 
It is worth noting that in our implementation, the number of model parameters is approximately 1.3M.

The proposed method was implemented using the PyTorch deep learning framework, and all model training was performed on NVIDIA RTX 3090 GPU hardware.
During training, a \text{batch\_size} of 512 was used, and the AdamW optimizer was chosen, which is well suited for HAR data.
The hyperparameters related to domain weights updating, specifically the step size $\eta$ and smoothing coefficient $c$, are set to 0.001 and 0.01, respectively.
For the reliability and comparability of the experiments, we used the CrossHAR \cite{hong2024crosshar} target model, strictly following the parameter settings of its original paper to ensure maximum consistency and rigor, thereby ensuring the reproducibility and scientific validity of the results.

\subsection{Performance on Multiple Source Datasets}

This paper studies the domain shift of jointly training models on different HAR datasets and the impact of the training data composition on the model performance. 
Therefore, we conducted cross-dataset transfer learning experiments, involving joint training on multiple labeled source datasets and evaluating model performance on unlabeled target datasets – a typical Domain Generalization setting. 
Specifically, we denote the HHAR dataset as H, Motion dataset as M, Shoaib dataset as S, and UCI dataset as U. 
The last column of Table \ref{tab2} shows the accuracy results of our proposed HAR-DoReMi method.


\begin{table*}[ht]
\centering
\small
\caption{Overall evaluation results (Accuracy) of multi-dataset training. (H, M, S, and U respectively represent HHAR, Motion, Shoaib, and UCI datasets. The \textbf{bold} and $\underline{\text{underline}}$ represent the best and second-best results.)}
\label{tab2}
\renewcommand{\arraystretch}{1.3} 
\begin{tabular}{@{}l *{10}{S[table-format=2.2]}@{}}
\toprule
\multicolumn{1}{c}{Model} & {\makecell{MLP}} & {\makecell{CNN}} & {\makecell{ResNet}} & {\makecell{LSTM}} & {\makecell{TSFCN}} & {\makecell{LIMUBE\\[0ex]RT}} & {\makecell{Contra\\[0ex]TSC}} & {\makecell{SDMix}} & {\makecell{Cross\\[0ex]HAR}} & \multicolumn{1}{c}{\textbf{\makecell{HAR-DoRe\\[0ex]Mi}}} \\
\midrule
HMS $\rightarrow$ UCI & 67.43 & 67.19 & 65.87 & 70.19 & 66.77 & 61.59 & 54.30 & 75.72  & $\underline{88.68}$ & \textbf{90.57} \\
HMU $\rightarrow$ Shoaib & 64.54 & 63.09 & 66.29 & 38.98 & 67.71 & 58.00 & 49.00 & 70.09  & $\underline{73.67}$ & \textbf{81.21} \\
HSU $\rightarrow$ Motion & 70.12 & 77.15 & 59.28 & 54.57 & 61.21 & 63.70 & 52.24 & 66.17  & $\underline{78.26}$ & \textbf{86.42} \\
MSU $\rightarrow$ HHAR & 61.47 & 58.71 & 56.75 & 42.97 & 51.08 & 63.26 & 43.36 & 74.44  & $\underline{76.19}$  & $\textbf{84.62}$ \\
\bottomrule
\end{tabular}
\end{table*}

\begin{figure*}[ht]
\centering
\includegraphics[width=\linewidth]{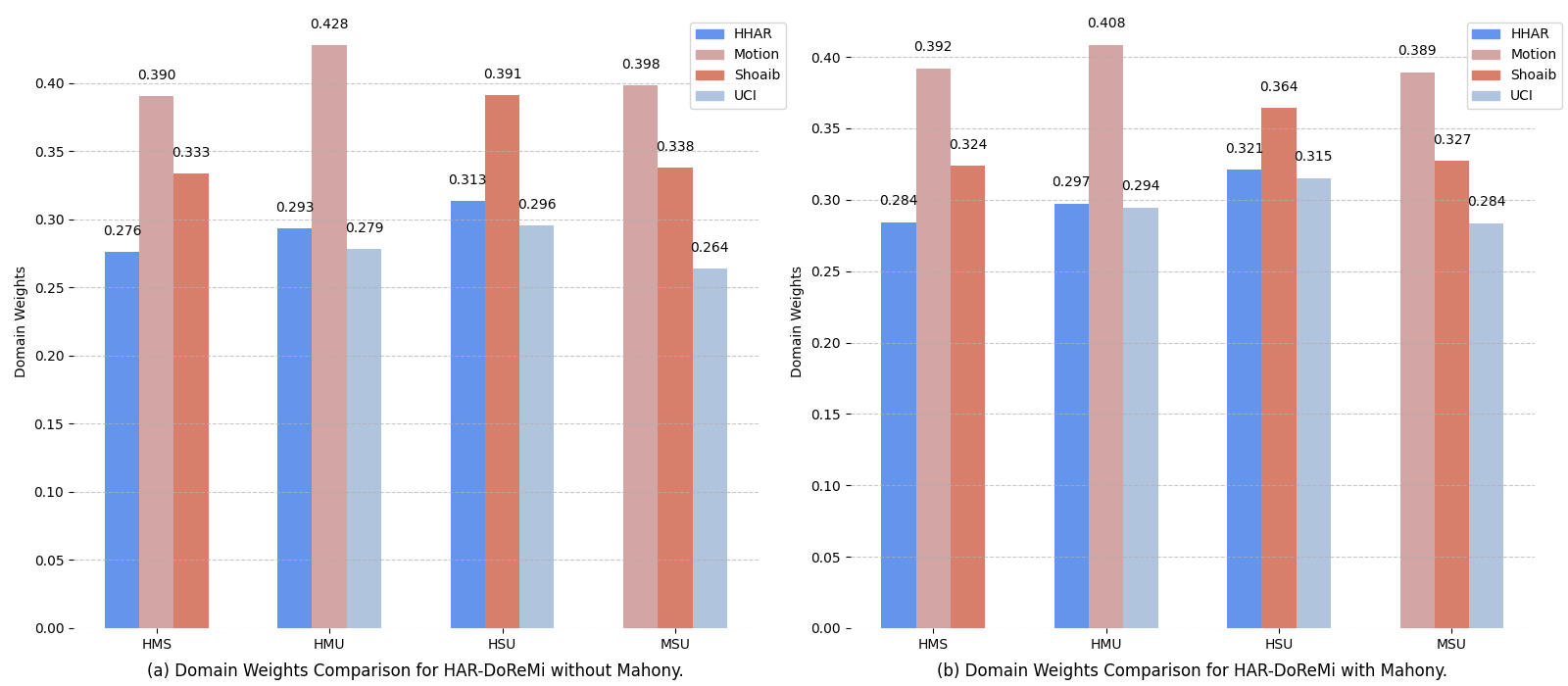}
\caption{Comparison of Domain Weights Obtained by HAR-DoReMi with and without Mahony Algorithm. (a) The domain weights derived from HAR-DoReMi without utilizing the Mahony algorithm for data transformation, reflect the initial importance and influence of the original dataset during training. (b)The domain weights obtained by HAR-DoReMi with the integration of the Mahony algorithm for data transformation, demonstrate the influence of Mahony algorithm on the weight distribution of the dataset during training. The results in the figure are all obtained at 1000 steps of training.
}
\label{fig_3}
\end{figure*}

Notably, in three transfer tasks, $\mathrm{HMS}\to\mathrm{UCI}$, $\mathrm{HMU}\to\mathrm{Shoaib}$, and $\mathrm{MSU}\to\mathrm{HHAR}$, HAR-DoReMi achieves excellent performance using only about 50\% of the training data used by CrossHAR.
In the $\mathrm{HSU}\to\mathrm{Motion}$ transfer task,the data usage is significantly reduced to about 30\% of CrossHAR.
Despite the significant reduction in data usage, our proposed method significantly outperforms the current state-of-the-art (SOTA) CrossHAR model in terms of accuracy on all transfer tasks, as shown in the results in Table \ref{tab2}. 
This provides convincing evidence for the effectiveness of our data mixture optimization strategy.
HAR-DoReMi achieves an average accuracy improvement of 6.51\% over the current SOTA methods in these four transfer tasks.
This method effectively alleviates the adverse effects of data distribution differences in heterogeneous HAR datasets in joint training through data mixture optimization, thereby significantly improving the model generalization ability.


In addition, we assume that there is a negative correlation between excess loss and model performance within a domain. Since the domain weight intuitively represents the excess loss, it can be considered an effective indicator of the difficulty of domain training. The domain weight during training reflects the attention of the model to the dataset. The higher the domain weight, the more difficult it is to train, requiring the model to allocate more learning resources to effectively capture the underlying patterns and features.

Fig. \ref{fig_3} shows the domain weights distribution of each dataset in different transfer learning scenarios. As shown in Fig. \ref{fig_3}, regardless of the application of the Mahony algorithm, the domain weights distribution shows significant differences between datasets. This strongly confirms the inherent differences in training difficulty between different datasets, verifying our previous hypothesis.
For example, in the $\mathrm{HMS}\to\mathrm{UCI}$ transfer task, the domain weights of the Motion dataset is relatively high, indicating that it is more difficult to train. This may be because the data distribution of the Motion dataset is very different from other datasets, making it difficult for the model to effectively learn its features.

In summary, the experimental results confirm the effectiveness of our data mixture optimization strategy and Mahony fusion algorithm, explaining the superior performance of HAR-DoReMi with low data usage.

\subsection{Performance on Shoaib Multi-sensor Dataset}

Existing HAR datasets exhibit significant differences in the number and location of sensors, which poses a great challenge to generalization across datasets.
To address the differences in feature dimensions due to sensor device heterogeneity and improve the applicability of models in datasets with different sensor configurations, this study decomposes multi-sensor data into separate sensor data streams.
In addition, in order to study the specific impact of sensor placement on model performance, we selected the Shoaib dataset containing multi-placed sensor data for detailed experimental analysis.

Since the Shoaib dataset explicitly records the sensor placements, we divide it into five subsets based on the sensor placements: \text{Shoaib\_LeftPocket} (\text{Shoaib\_LP}), \text{Shoaib\_RightPocket} (\text{Shoaib\_RP}), \text{Shoaib\_Wrist} (\text{Shoaib\_W}), \text{Shoaib\_UpperArm} (\text{Shoaib\_UA}), and \text{Shoaib\_Belt} (\text{Shoaib\_B}). Splitting the Shoaib dataset into multiple single-sensor datasets enables us to simulate different sensor configurations and evaluate the performance of HAR-DoReMi using such data.


\begin{table*}[ht]
\centering
\small
\caption{Overall evaluation results (Accuracy) of multi-sensor dataset training. (The \textbf{bold} and $\underline{\text{underline}}$ represent the best and second-best results.)}
\label{tab3}
\renewcommand{\arraystretch}{1.3} 
\sisetup{table-format=2.2}
\begin{tabular}{
    l | 
    l
    S[table-format=2.2]
    S[table-format=2.2]
    S[table-format=2.2]
    S[table-format=2.2]
    S[table-format=2.2]
    S[table-format=2.2]
    S[table-format=2.2]
    S[table-format=2.2] 
    S[table-format=2.2]
    >{\bfseries}c
}
\toprule
{\makecell[c]{Source}} & {\makecell[c]{Target}} & {\makecell[c]{MLP}} & {\makecell[c]{CNN}} & {\makecell[c]{ResNet}} & {\makecell[c]{LSTM}} & {\makecell[c]{TSFCN}} & {\makecell[c]{LIMUBE\\[0ex]RT}} & {\makecell[c]{Contra\\[0ex]TSC}} & {\makecell[c]{SDMix}} & {\makecell[c]{Cross\\[0ex]HAR}} & {\makecell[c]{HAR-DoRe\\[0ex]Mi}} \\
\midrule
\multirow{3}{*}{Shoaib}
 & UCI      & 60.64 & 64.90 & 66.53 & 59.50 & 70.37 & 53.00 & 55.36 & $\underline{71.77}$ & 67.81 & \textbf{82.75} \\
 & Motion   & 58.59 & 72.05 & 59.18 & $\underline{73.78}$ & 60.03 & 69.50 & 53.55 & 65.07 & 72.18 & \textbf{79.77} \\
 & HHAR     & 50.18 & 59.50 & 58.56 & 57.45 & 60.57 & 53.00 & 42.86 & 58.31 & $\underline{67.03}$ & \textbf{71.60} \\
\bottomrule
\end{tabular}
\end{table*}

\begin{figure}
\centering
\includegraphics[width=\linewidth]{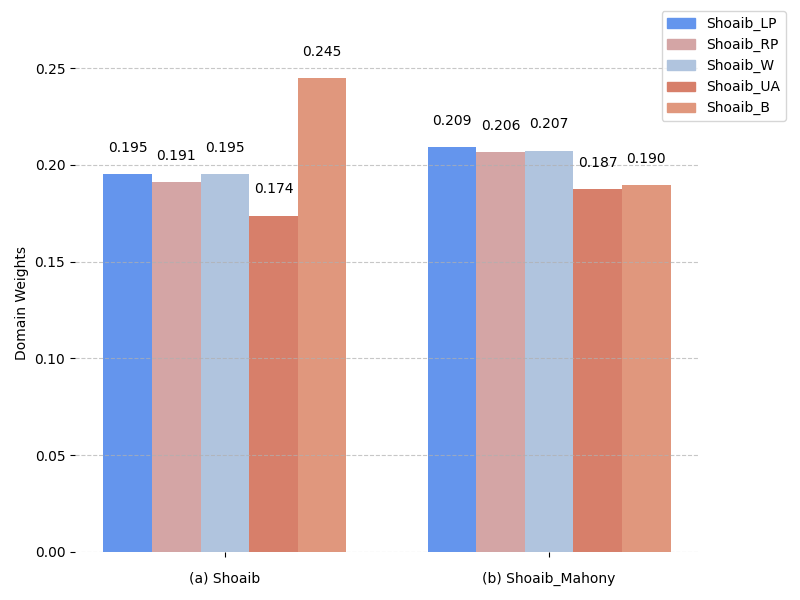}
\caption{Domain Weights Comparison for Shoaib Dataset Sensor Placement Subsets. The figure presents domain weights from HAR-DoReMi training on Shoaib dataset subsets (partitioned by sensor location) at 1000 training steps. (a) Domain weights without Mahony algorithm data transformation. (b) Domain weights with Mahony algorithm data transformation.}
\label{fig_4}
\end{figure}

We used the partitioned Shoaib dataset as the source domain and conducted transfer learning experiments on other datasets. 
Table \ref{tab3} shows the overall evaluation results (accuracy) of HAR-DoReMi on multi-sensor datasets. 
The results show that the HAR-DoReMi method achieves significant performance improvements in all transfer tasks. 
The experimental results further show that the average accuracy of this method is about 7.18\% higher than the current state-of-the-art model. 
Notably, HAR-DoReMi continues to perform well on multi-sensor datasets, achieving even greater performance improvements.

Fig. \ref{fig_4} shows the distribution of domain weights for different sensor placement subsets in the Shoaib dataset of the HAR-DoReMi framework.
As shown in Fig. \ref{fig_4}, regardless of whether the Mahony algorithm is used, the domain weights show certain differences on different sensor placement subsets, which indicates that the training difficulty of sensor data from different placements is different.

In summary, the experiments on the Shoaib dataset analyzed the impact of sensor configuration data on model training. 
The results show that HAR-DoReMi can effectively process multi-sensor configuration data and achieve excellent performance.

\subsection{Effectiveness of Mahony Algorithm}

In addition to exploring joint training across different HAR datasets, we further study the performance of a single HAR dataset in cross-dataset training, with a particular focus on evaluating the role of Mahony algorithm in leveraging data pre-processing to improve cross-domain generalization capabilities.
We innovatively introduced the Mahony fusion algorithm originally derived from posture control into self-supervised HAR, aiming to effectively reduce data heterogeneity by unifying the data coordinate system across datasets, thereby significantly enhancing the generalization ability of the pre-training model across datasets.

\begin{figure*}[ht]
\centering
\includegraphics[width=\linewidth]{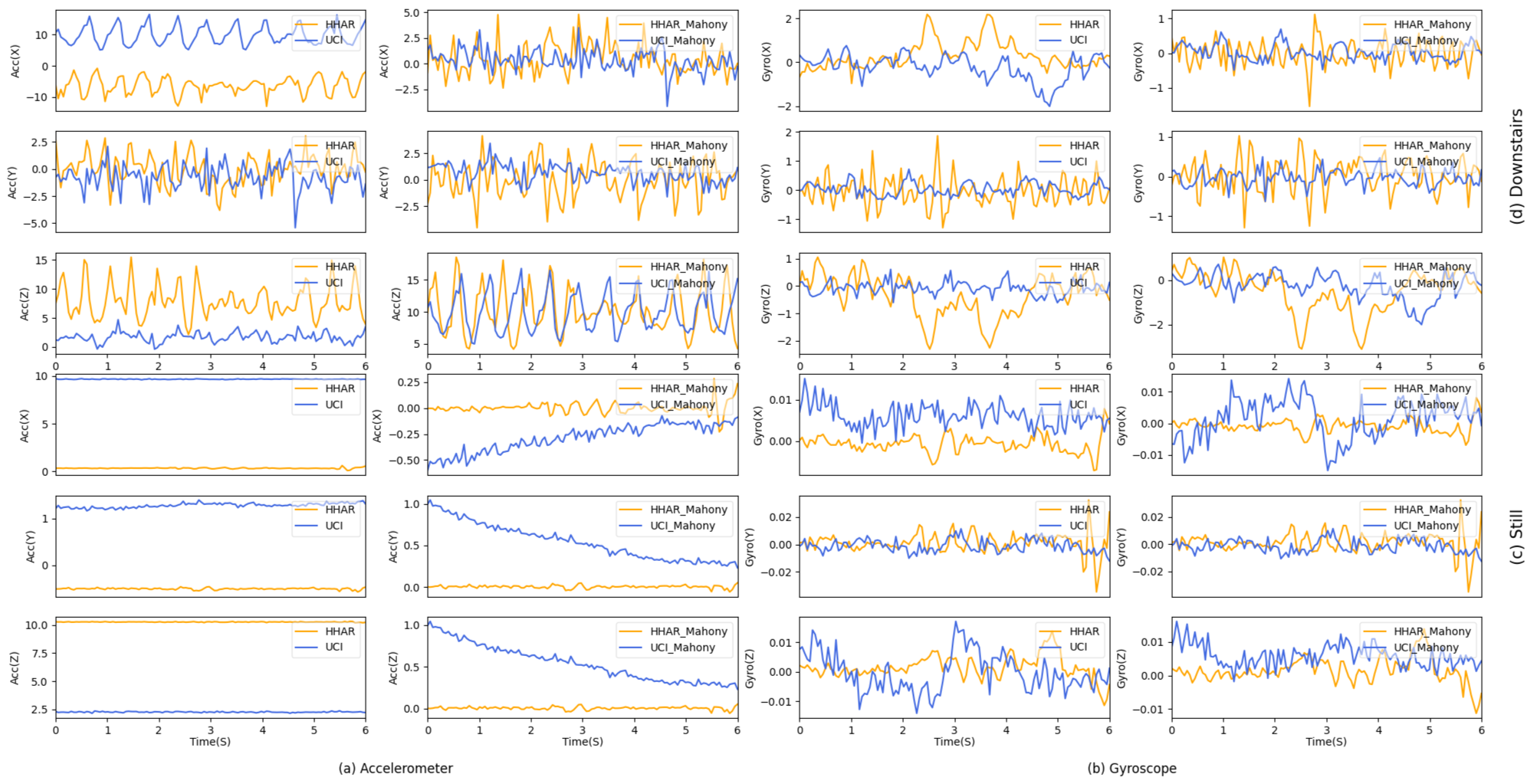}
\caption{Comparison Figure of the Mahony Algorithm's Alignment Effect on IMU Data for Identical Activities in the HAR Dataset. The upper half of the figure illustrates the application effect of the Mahony algorithm on the \textquotedblleft{}Downstairs\textquotedblright{} activity. The left side shows the raw data, while the right side presents the data processed by the Mahony algorithm. The lower half demonstrates the effect of the Mahony algorithm on the \textquotedblleft{}Still\textquotedblright{} activity. Similarly, the left side displays the raw data, and the right side shows the processed data.}
\label{fig_5}
\end{figure*}

To quantify the effectiveness of the Mahony algorithm, we used the CrossHAR architecture \cite{hong2024crosshar} and pre-processing the input data with Mahony algorithm.
Fig. \ref{fig_5} visually demonstrates the impact of Mahony algorithm on aligning IMU data of the same activity across datasets. Using \textquotedblleft{}Downstairs\textquotedblright{} and \textquotedblleft{}Still\textquotedblright{}as examples, Fig. \ref{fig_5} clearly compares the original data and the data processed by Mahony algorithm.
The original data shows clear differences due to sensor orientation and gravity.
However, Mahony algorithm processing makes the data of the same activity across datasets more consistent in waveform and amplitude, indicating that Mahony algorithm effectively unifies the sensor coordinate system, providing more consistent input for model training.
This coordinate alignment helps the model learn shared features across datasets, thereby improving generalization ability.
Moreover, Fig. \ref{fig_3} and \ref{fig_4} demonstrate a clear trend: the application of the Mahony algorithm results in a more homogeneous domain weights distribution for the datasets of different transfer tasks. It is evident that employing the Mahony algorithm brings the domain weights distributions of datasets in different transfer tasks closer together, indirectly demonstrating the algorithm's effectiveness in reducing data heterogeneity.

\begin{table*}[ht]
\centering
\small
\caption{Overall results (Accuracy) of the Mahony algorithm in cross-dataset setting. (The \textbf{bold} and $\underline{\text{underline}}$ represent the best and second-best results.)}
\label{tab4}
\sisetup{table-format=2.2}
\renewcommand{\arraystretch}{1.3} 
\begin{tabular}{
    l| 
    l
    S[table-format=2.2]
    S[table-format=2.2]
    S[table-format=2.2]
    S[table-format=2.2]
    S[table-format=2.2]
    S[table-format=2.2]
    S[table-format=2.2]
    S[table-format=2.2]
    S[table-format=2.2]
    >{\bfseries}c
}
\toprule
{\makecell[c]{Source}} & {\makecell[c]{Target}} & {\makecell[c]{MLP}} & {\makecell[c]{CNN}} & {\makecell[c]{ResNet}} & {\makecell[c]{LSTM}} & {\makecell[c]{TSFCN}} & {\makecell[c]{LIMUBE\\[0ex]RT}} & {\makecell[c]{Contra\\[0ex]TSC}} & {\makecell[c]{SDMix}} & {\makecell[c]{Cross\\[0ex]HAR}} & {\makecell[c]{CrossHAR\\[0ex]+Mahony}} \\
\midrule
\multirow{3}{*}{UCI} 
 & Shoaib   & 46.74 & 49.38 & 50.51 & 32.50 & 36.45 & 58.28 & 51.01 & 57.67 & $\underline{69.93}$ & \textbf{77.25} \\
 & Motion   & 53.08 & 43.68 & 52.49 & 64.75 & 46.19 & 68.94 & 52.29 & 66.18 & \textbf{87.90} & $\underline{80.21}$ \\
 & HHAR     & 45.94 & 56.46 & 51.21 & 68.15 & 44.81 & 69.44 & 42.86 & 61.10 & $\underline{77.27}$ & \textbf{85.26} \\
\midrule
\multirow{3}{*}{Shoaib}
 & UCI      & 60.64 & 64.90 & 66.53 & 59.50 & 70.37 & 53.00 & 55.36 & $\underline{71.77}$ & 67.81 & \textbf{78.45} \\
 & Motion   & 58.59 & 72.05 & 59.18 & $\underline{73.78}$ & 60.03 & 69.50 & 53.55 & 65.07 & 72.18 & \textbf{79.54} \\
 & HHAR     & 50.18 & 59.50 & 58.56 & 57.45 & 60.57 & 53.00 & 42.86 & 58.31 & $\underline{67.03}$ & \textbf{74.59} \\
\midrule
\multirow{3}{*}{Motion}
 & UCI      & 56.43 & 50.06 & 63.52 & 53.31 & 57.51 & 72.26 & 45.05 & 65.09 & $\underline{76.41}$ & \textbf{83.35} \\
 & Shoaib   & 57.21 & 56.92 & 60.37 & 56.82 & 55.74 & 65.72 & 43.21 & 59.93 & $\underline{70.89}$ & \textbf{82.53} \\
 & HHAR     & 52.37 & 43.12 & 53.49 & 43.70 & 49.62 & 64.63 & 42.86 & $\underline{74.09}$ & 65.12 & \textbf{79.81} \\
\midrule
\multirow{3}{*}{HHAR}
 & UCI      & 54.51 & 48.85 & 48.02 & 40.50 & 48.02 & 67.04 & 55.54 & 54.49 & $\underline{91.52}$ & \textbf{92.71} \\
 & Shoaib   & 53.89 & 24.39 & 44.88 & 19.45 & 41.14 & 53.56 & 26.24 & 57.49 & $\underline{61.08}$ & \textbf{72.36} \\
 & Motion   & 60.28 & 60.86 & 51.64 & 37.11 & 64.38 & 57.04 & 65.87 & 61.14 & $\underline{75.17}$ & \textbf{81.02} \\
\bottomrule
\end{tabular}
\end{table*}

To more quantitatively evaluate the impact of Mahony algorithm on cross-domain generalization, we designed a transfer learning experiment: training the model on one dataset and evaluating it on another dataset.
Table \ref{tab4} shows the results.
It can be seen that using the Mahony algorithm for data pre-processing significantly improves model performance in all transfer tasks, which strongly supports the effectiveness of Mahony algorithm in reducing data heterogeneity.
These findings clearly demonstrate the positive role of Mahony algorithm in improving cross-dataset training performance, effectively alleviating data heterogeneity, and significantly improving the generalization ability of pre-training models across datasets.

\subsection{Ablation Experiment}

This section details the ablation experiments aimed at analyzing the performance contributions of individual HAR-DoReMi components and verifying the effectiveness of our proposed approach.
For comparison, we distinguish between versions of HAR-DoReMi with and without Mahony algorithm pre-processing. 
All experiments are conducted on four transfer tasks: $\mathrm{HMS}\to\mathrm{UCI}$, $\mathrm{HMU}\to\mathrm{Shoaib}$, $\mathrm{HSU}\to\mathrm{Motion}$, and $\mathrm{MSU}\to\mathrm{HHAR}$, consistently using the CrossHAR architecture as the target model.
To clarify, the Mahony algorithm is used as a data pre-processing step to transform the input data and the model architecture remains CrossHAR throughout.


\begin{table*}[ht]
\centering
\caption{Overall Results (Accuracy) of Different Components in Multi-Dataset Setting. 
(The \textbf{bold} and \underline{underline} represent the best and second-best results, 
and the values in brackets are the percentage of the used data in the CrossHAR baseline model.)}
\label{tab5}
\small
\renewcommand{\arraystretch}{1.3} 
\begin{tabular}{@{} l *{4}{c} @{}}
\toprule
\multicolumn{1}{c}{Model} &
\multicolumn{1}{c}{\thead{CrossHAR}} &
\multicolumn{1}{c}{\thead{CrossHAR \\ + Mahony}} &
\multicolumn{1}{c}{\thead{HAR-DoReMi}} &
\multicolumn{1}{c}{\thead{HAR-DoReMi \\ + Mahony}} \\
\midrule
HMS $\rightarrow$ UCI & 88.68 \text{\footnotesize (100\%)} & \textbf{91.02} \text{\footnotesize (100\%)} & 89.21 \text{\footnotesize (53.74\%)} & \underline{90.57} \text{\footnotesize (53.53\%)} \\
HMU $\rightarrow$ Shoaib & 73.67 \text{\footnotesize (100\%)} & \underline{79.88} \text{\footnotesize (100\%)} & 79.04 \text{\footnotesize (44.28\%)} & \textbf{81.21} \text{\footnotesize (41.91\%)} \\
HSU $\rightarrow$ Motion & 78.26 \text{\footnotesize (100\%)} & \underline{80.39} \text{\footnotesize (100\%)} & 80.14 \text{\footnotesize (33.27\%)} & \textbf{86.42} \text{\footnotesize (31.21\%)} \\
MSU $\rightarrow$ HHAR & 76.19 \text{\footnotesize (100\%)} & 78.93 \text{\footnotesize (100\%)} & \textbf{89.46} \text{\footnotesize (48.06\%)} & \underline{84.62} \text{\footnotesize (44.75\%)} \\
\bottomrule
\end{tabular}
\end{table*}

In all experiments, HAR-DoReMi and HAR-DoReMi + Mahony use about 30\% to 50\% of the data of CrossHAR and CrossHAR + Mahony models, demonstrating the data efficiency of our method.
Table \ref{tab5} shows the results of the ablation study. From this table, we can conclude that:

(1) Data efficiency:
HAR-DoReMi and HAR-DoReMi + Mahony perform comparable to or better than CrossHAR and CrossHAR + Mahony on multiple transfer tasks. 
For example, on $\mathrm{MSU}\to\mathrm{HHAR}$, HAR-DoReMi and HAR-DoReMi + Mahony achieve 89.46\% and 84.62\% accuracy, respectively, significantly outperforming CrossHAR (76.19\%) and CrossHAR + Mahony (78.93\%). 
This clearly demonstrates the effectiveness of our data mixture optimization strategy in improving data efficiency, achieving the same or better results with less data.

(2) Mahony Algorithm Contribution:
Table \ref{tab5} compares the versions with and without Mahony algorithm, and the results show that using Mahony algorithm improves performance in most transfer tasks, especially in the $\mathrm{HMU}\to\mathrm{Shoaib}$ transfer task. 
This shows that the Mahony algorithm effectively reduces the data heterogeneity by aligning the sensor coordinate system, providing more consistent input for model training, thereby enhancing the model generalization ability.

(3) HAR-DoReMi + Mahony Combination Advantages: 
The HAR-DoReMi + Mahony combined model showed excellent performance in all transfer tasks, and achieved a peak accuracy of 86.42\% in the $\mathrm{HSU}\to\mathrm{Motion}$ transfer task. 
This reflects the effective complementarity of the HAR-DoReMi and Mahony algorithm: Mahony algorithm reduces data heterogeneity and improves the input of HAR-DoReMi, while HAR-DoReMi further improves the model generalization ability through data mixture optimization.

In summary, the ablation study results clearly demonstrate the performance contributions of the HAR-DoReMi approach and its components. 
The HAR-DoReMi + Mahony combination achieves the best balance of data efficiency, performance, and generalization, confirming the effectiveness and superiority of our proposed approach for cross-dataset HAR.

\section{DISSCUSION}

Although the HAR-DoReMi framework inspired by LLMs and based on DoReMi has significantly advanced cross-dataset HAR through effective data mixture optimization, it is important to recognize its inherent limitations and areas for future improvements.

(1) Applicability and Limitations of the Mahony Pose Fusion Algorithm: 
To address the problem of IMU sensor orientation heterogeneity, we innovatively incorporate the Mahony algorithm originally designed for pose estimation into self-supervised HAR pre-training.
Experiments show that Mahony algorithm has a positive impact on cross-dataset generalization.
However, as a filter-based complementary pose estimation method, the performance of Mahony algorithm is also susceptible to sensor noise, dynamic environment, and initial conditions.
In the case of extreme motion or poor sensor data quality, the pose estimation accuracy of Mahony algorithm may degrade, which may affect the performance of the HAR-DoReMi framework.
Future work can explore more advanced and robust pose estimation algorithms, aiming to improve the robustness and accuracy of data pre-processing.

(2) Theoretical Analysis and Generalization Bounds: 
Although the experiments validate the effectiveness of HAR-DoReMi, our theoretical understanding of data mixture optimization is still limited. 
For example, while the domain weights optimization based on the Group DRO algorithm can minimize the worst domain loss, its optimality for cross-dataset HAR is uncertain.
The impact of data mixture ratio on generalization and the connection between domain weights and generalization ability also need further study. 
These theoretical issues deserve further exploration and resolution.
In addition, further research is needed to study the performance of HAR-DoReMi on more diverse and larger HAR datasets to more comprehensively evaluate its generalization and applicability.


In summary, HAR-DoReMi provides a novel and effective data mixture optimization strategy for cross-dataset HAR. 
By recognizing and actively mitigating the above limitations and continuously improving them, we expect HAR-DoReMi and related technologies to play a more important role in ubiquitous computing and human-computer interaction. 
This will promote the wider application of HAR technology in fields such as health monitoring, smart homes, and motion analysis.

\section{CONCLUSION}

In this paper, we have applied the data mixture optimization to Human Activity Recognition (HAR) pre-training and proposed the novel HAR-DoReMi framework as an effective solution for cross-dataset generalization.
Inspired by large language models (LLMs) and tailored for HAR, HAR-DoReMi significantly improves the data efficiency and cross-dataset generalization performance of pre-training HAR models.
Experimental validation shows that HAR-DoReMi outperformed the current state-of-the-art models even with less data, highlighting its data efficiency and strong generalization ability.
HAR-DoReMi is expected to inspire future research and promote the further development of HAR technology in ubiquitous computing and human-centered artificial intelligence.


\bibliographystyle{unsrt}
\bibliography{sample-base}

\vfill

\end{document}